\documentclass[letterpaper, 10 pt, conference]{ieeeconf}  

\IEEEoverridecommandlockouts                              

\usepackage{graphicx} 
\usepackage{amsmath} 

\usepackage{array}
\usepackage{ulem}

\usepackage{url}

\title{\LARGE \bf
Robotic Defect Inspection with Visual and Tactile Perception for Large-scale Components
}

\author{Arpit Agarwal$^{1}$, Abhiroop Ajith$^{2}$, Chengtao Wen$^{2}$,   Veniamin Stryzheus$^{3}$, Brian Miller$^{3}$, \\Matthew Chen$^{3}$, Micah K. Johnson$^{4}$, Jose Luis Susa Rincon$^{2}$, Justinian Rosca$^{2}$ and Wenzhen Yuan$^{1}$
\thanks{$^{1}$Arpit Agarwal and Wenzhen Yuan are with the Robotics Institute, Carnegie Mellon University
        {\tt\small \{arpita1, wenzheny\}@andrew.cmu.edu}}
\thanks{$^{2}$Chengtao Wen, Abhiroop Ajith, Jose Luis Susa Rincon, and Justinian Rosca are with Siemens Corporations 
    {\tt\small \{chengtao.wen, abhiroop.ajith, jose.susa\_rincon, justinian.rosca\}@siemens.com}}
\thanks{$^{3}$Matthew Chen, Veniamin Stryzheus and Brian Miller are with Boeing 
    {\tt\small \{matthew.j.chen, veniamin.v.stryzheus, brian.t.miller2\}@boeing.com}}
\thanks{$^{4}$Micah K. Johnson is with GelSight Inc. {\tt\small kimo@gelsight.com}}
}    

\usepackage[colorinlistoftodos, disable]{todonotes}

\newcommand{\wenzhen}[1]{\todo[inline,color=red!40]{#1}}

\newcommand{\boeing}[1]{\todo[inline,color=yellow!40]{#1}}
\begin{document}

\maketitle
\thispagestyle{empty}
\pagestyle{empty}
\begin{abstract}
In manufacturing processes, surface inspection is a key requirement for quality assessment and damage localization. Due to this, automated surface anomaly detection has become a promising area of research in various industrial inspection systems. A particular challenge in industries with large-scale components, like aircraft and heavy machinery, is inspecting large parts with very small defect dimensions. Moreover, these parts can be of curved shapes.  
To address this challenge, we present a 2-stage multi-modal inspection pipeline with visual and tactile sensing. Our approach combines the best of both visual and tactile sensing by identifying and localizing defects using a global view (vision) and using the localized area for tactile scanning for identifying remaining defects. 
To benchmark our approach, we propose a novel real-world dataset with multiple metallic defect types per image, collected in the production environments on real aerospace manufacturing parts, as well as online robot experiments in two environments. Our approach is able to identify 85\% defects using Stage I and identify 100\% defects after Stage II. The dataset is publicly available at \url{https://zenodo.org/record/8327713}.

\end{abstract}

\section{Introduction}
Various large-scale manufacturing machinery and industries with large metal parts like aircraft components, experience various internal and external factors such as vibration, foreign objects debris, high temperature, friction, and corrosion. This can lead to fatigue or even part failure. 
Hence, to ensure safe operation, each industry requires surface inspection. For example, in the aircraft industry airplanes are inspected every 100 hours\cite{aopa_2016}, according to Federal Aviation Administration(FAA) rules. The periodic inspection could extend the lifetime of the parts. However, human visual and touch inspection still accounts for more than 90\% of inspection checks\cite{nicklesdescriptive}.

There is a significant interest in automating the surface defect detection process, as it allows for fast, repeatable, and cost-effective detection, as compared to the human expert inspection process.
Surface defect detection on industrial parts is a fast-growing  market \cite{marketsandmarkets_2020}. 
Nowadays, more and more inspection systems use vision-based techniques combined with Deep Learning for defect detection\cite{roth2022towards} \cite{jahanshahi2013innovative}. 
\boeing{current requirement is a fingernail could catch - 0.01mm}
However, aerospace and spacecraft industries have different inspection requirements - they have large metal parts which need to be scanned and the dimensions of the defects can be as small as 0.01mm.  

Instead of relying on a vision-only system, we propose a visuotactile 2-stage pipeline for surface defect detection. Our method combines the advantages of both vision and tactile sensing and avoids their limitations: vision has high prediction speed and can cover large surface area, but typically attains low accuracy since the visual appearance of defects can be influenced by many sources of noise; contrarily, high-resolution tactile sensing, give high accuracy but has low speed because of the small coverage area in a single scan. The first stage of our pipeline uses an RGB camera to collect an image of a segment of the specimen and uses deep learning to identify potential defect regions. The regions with low defect confidence are passed onto the second stage of the pipeline which leverages a high-resolution vision-based tactile sensor, the GelSight Mobile, for taking a tactile scan. This tactile data is used to identify and classify the surface defect. This approach allows the scanning of large surfaces for small anomalies efficiently. We implemented the whole system on a robot arm, to allow for inspection in a production environment. Using our method, we are able to identify defects 100\% of the time in a fraction of the time as compared to the tactile-only  approach and more accurately than the vision-only approach.

We make 3 specific contributions in this work
\begin{itemize}
    \item We introduce the first aerospace defect detection dataset containing metallic surfaces with multiple defects in a single image
    \item We propose a 2-stage defect detection approach using visuotactile sensing
    \item We integrate our detection approach into a prototype system on an industrial robot arm
\end{itemize}
We introduce the dataset and dataset collection details in Section (\ref{sec:dataset}), the visuotactile detection approach in Section (\ref{sec:approach}), and the integrated robot system for runtime defect detection in Section \ref{sec:robotsystem}. Using our approach, we are able to achieve perfect recall in 70x less inspection as compared to the tactile-only approach. We successfully integrate our detection system in 2 separate environments(different arms, different illumination conditions, and different panels). The proposed techniques are widely applicable to various industries with large-scale components like ship hull inspection and heavy machinery.   

\section{Related Work}

This section surveys works that present novel defect detection techniques as well as works that propose datasets with industrial defects.

\textit{Defect detection methods}: This section covers various surface inspection techniques using different sensing techniques. 
In \cite{jahanshahi2013innovative}, authors used a depth camera to create a 3D reconstruction of the part under inspection, computer vision techniques for segmenting cracks, and machine learning for classifying them into defect vs non-defect patterns. In the aerospace industry, the most common type of part is metallic and very reflective. As noted in \cite{weng2020multi}, commercial depth sensors exhibit poor accuracy when sensing metallic parts. In \cite{zou2018deepcrack}, authors train a custom deep CNN for automatic crack detection on concrete surfaces. Their approach gives a 0.87 F-measure value on the CrackLS315 dataset. In \cite{jiang2021vision}, authors similarly used a CNN and a vision-based tactile sensor for crack profile estimation. However, it is unclear how to extend the approach to images containing multiple kinds of defects that are not scratches.    
\cite{palermo2021multi} is the closest to our work. They propose a 2-stage visuotactile pipeline 
targeted only to crack detection. They used 3500 images to train an object detector and used an optical fiber-based tactile-proximity sensor for assessing cracks. However, their method is tested on a toy dataset using 3D-printed parts containing cracks in a lab setting. Their dataset contains a single large crack across the image on a non-metallic surface. We have integrated our detection pipeline in a production setting and show results on real aerospace parts. Moreover, we require an order of magnitude less data than their work. 

\textit{Metal defect datasets}: In this section, we cover datasets that target defect detection in industrial parts and manufacturing processes. 
MVTec dataset \cite{bergmann2019mvtec} introduced a challenging dataset for unsupervised anomaly detection. The dataset contains RGB images of various small to medium manufactured parts like carpets, leather, metal nut, cable, hazelnut, etc. 
However, each image contains only a single type of anomaly. In comparison, our dataset contains multiple defects in a single image and can be very small(less than 2\% of pixels in the image). 
The magnetic tile dataset\cite{magnetictileHuang2020surface} contains 1344 grayscale images along with a segmentation mask for each image.
They provide a segmentation mask for each image. The dataset is targeted towards industrial parts (flat metallic sheets),  which are challenging to image, similar to our case. However, the parts considered in the dataset are flat and have consistent illumination across the tile plane. This illumination setting is hard to replicate for aerospace parts, which can be curved and have a significant variation in color across the metallic part. 

\section{Boeing-CMU multi-defect dataset}\label{sec:dataset}
We introduce a novel dataset of surface defect detection for aerospace metal parts. This dataset is used to test our defect detection algorithm in an offline setting. Our dataset contains 184 RGB images with bounding box annotations in Pascal VOC format\cite{everingham2010pascal} for each image. Each RGB image contains multiple defects. The defects are manually made on the parts by experts from Boeing with a process similar to the real defects in production, and they are a more challenging inspection cases since the defect density is higher than real parts in production.  
Each bounding box contains the location and class of the defect. 
This dataset contains 3 kinds of defects - \textit{scratches}, \textit{drill runs}, and \textit{gouges}. Figure \ref{fig:boeingdefects} illustrates the defects by showing its RGB images, GelSight tactile image, and depth profile along the defect, respectively. The standard definition of the defects is given in terms of depth and width of surface geometry, as marked in the \textit{Heightfield} in Figure \ref{fig:boeingdefects}. Table \ref{tab:defectoccurences} shows the breakdown of the number of defects in our dataset. 

\begin{figure}
    \centering
    \includegraphics[width=\columnwidth]{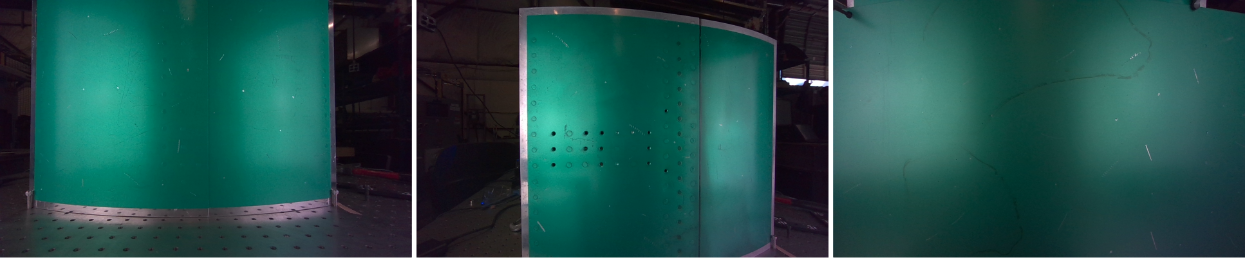}
    \caption{\textbf{Dataset Illustration}: It contains RGB images of aircraft parts from Boeing. Each panel is curved with 3 sizes 40in $\times$ 40in,  56in $\times$ 18in, and 94in $\times$ 20in. For each image, we have bounding box annotations made by industry inspectors.}
    \label{fig:datasetex}
\end{figure}

\begin{figure}
    \centering
    \includegraphics[width=0.8\columnwidth]{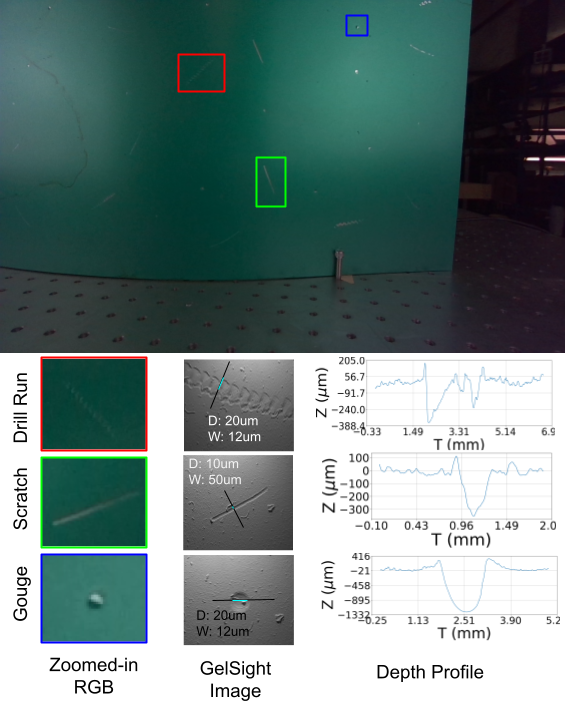}
    \caption{\textbf{Dataset defect description}: The top image shows an RGB image and 3 types of defect. The bottom 3 rows show(left-to-right) zoomed-in RGB image, heightfield of the anomalous region, and detrended depth profile.}
    \label{fig:boeingdefects}
\end{figure}

\begin{figure}
    \centering
    \includegraphics[width=\columnwidth]{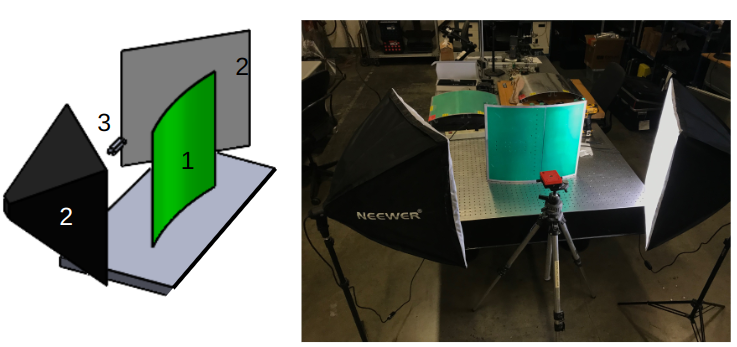}
    \caption{\textbf{Dataset capture setup}: Left image contains - (1) Position for metal panel placement; (2) Neewer 24 in $\times$ 24 in soft boxes lights with 700 Watt, 5500K CFL Light Bulbs;  (3) RealSense D435 camera. On the right, we show the real setup used to collect images for our dataset.}
    \label{fig:boeingcapture}
\end{figure}

The dataset was collected at the Boeing lab with an Intel RealSense D455 camera at a resolution of 1280 $\times$ 800. The full setup is shown in Figure \ref{fig:boeingcapture}. We placed soft boxes (bulb with a diffuser cloth in front) at an angle of 45$^{\circ}$ along the vertical axis on either side of the camera. This illumination setting allows us to capture images of metallic curved panels without over-saturation or under-exposure in any part of the image. For the dataset, we used 18 curved metal (approximate radius of curvature 26.5 inch) panels - 2 panels of dimension 40 inch $\times$ 40 inch, 15 panels of dimensions 56 inch $\times$ 38 inch, and 1 panel of dimension 94in $\times$ 20in. We collect 9 images at different locations per panel to cover the whole panel. Each panel is a piece of an aircraft with fasteners, a support structure underneath, and a green temporary protective coating. All the images were manually labeled by Boeing personnel using LabelImg\footnote{https://github.com/heartexlabs/labelImg}, a graphical image annotation tool. Figure \ref{fig:datasetex} shows some illustrative images in the datasets. One noticeable feature is the presence of significant variation in the surface color. This is due to the surface being curved and metallic in appearance. 

\begin{table}
    \centering
    \caption{Description of defects in our dataset}
    \begin{tabular}{m{4em}||m{1.2cm}|m{1.2cm}||m{4em}|m{4.5em}}
        \textbf{Defect Type} & \textbf{Minimum depth(mm)} & \textbf{Minimum width(mm)} & \textbf{Number of occurrences} & \textbf{Number of physical defects}\\
        \hline
        Drill Run & 0.02 & 0.012 & 927 & 122\\
        Scratch & 0.01 & 0.051 & 1648 & 221\\
        Gouge & 0.02 & 0.012 & 2431 & 312
    \end{tabular}
    \label{tab:defectoccurences}
\end{table}




\textbf{Tactile dataset}: We collected tactile data using a GelSight Mobile 0.5X\cite{gelsightprobe}, a high-resolution vision-based tactile sensor with 6.9$\mu$m resolution in x-y direction and 4$\mu$m in the z-direction. 
We manually pressed the sensor on the probable defect location. We collected 59 scans from 1 Boeing panel, containing - 17 scratches, 14 gouges, 
 and 18 drill runs. We also collect 10 no-defect cases.\wenzhen{You should mention that some no-defect cases contains significant features such as the fasteners and the edges. Also it's nice to show some of those cases somewhere, e.g. combine that with Fig 11.} Each tactile scan is manually labeled with a class label.  

\section{Multi-modal defect detection method and setup}\label{sec:approach}
\begin{figure*}[h]
    \centering
    \includegraphics[width=0.8\textwidth]{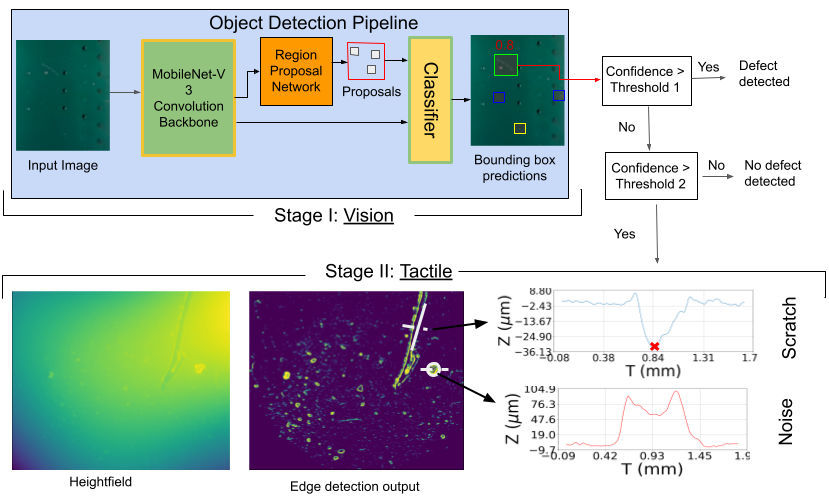}
    \caption{\textbf{Detection Overview}: Our approach consists of 2 stages A) Vision stage uses Deep Learning based bounding box detector for identifying defects in the RGB image from the global view. B) Based on the confidence threshold we identify defects or send them to stage 2. C) Tactile stage uses the high-resolution heightfield extracted from GelSight and inspects the depth profile of anomalous regions to identify the type of defect.} 
    \label{fig:overview}
\end{figure*}

Figure \ref{fig:overview} shows the proposed pipeline for surface defect detection and classification based on visual and tactile sensing. Our 2-stage pipeline uses RGB images for identifying defect regions with a confidence value. We delegate bounding boxes with low confidence scores to the second stage and use high-resolution tactile images for identifying the defect. In the following section, we provide details about each stage. 

\subsection{Stage I: Vision-based defect detection}
\label{sec:visiondetect}
The first stage uses an RGB camera to scan the surface and predict defects. We used a Faster Region-based Convolutional Neural Network(Faster R-CNN)\cite{fasterrcnn} with MobileNet-v3 FPN backbone \cite{mobilenetv3}. The neural network architecture was chosen based on empirical observation. 
The model was pretrained on Common Objects in Context (COCO) dataset \cite{coco}. We fine-tune the last 3 backbone layers, regression, and classification models after feature prediction. Note, the model can be used with images of any size without resizing, at both train and test time, as it is fully convolutional. The neural network model predicts multiple bounding boxes per image. Each bounding box contains the coordinates of the rectangle region in the camera coordinate frame, defect class, and confidence score for that class. 
At test time, we predict bounding boxes with a confidence score higher than 0.7 as surface defects with certainty and shortlist those with scores between 0.1 and 0.7 to delegate to the next stage of the pipeline. These threshold choices provide a good trade-off between detection in stage I and proposing candidates with minimal false positives for stage II.

\begin{figure}
    \centering
    \includegraphics[width=\columnwidth]{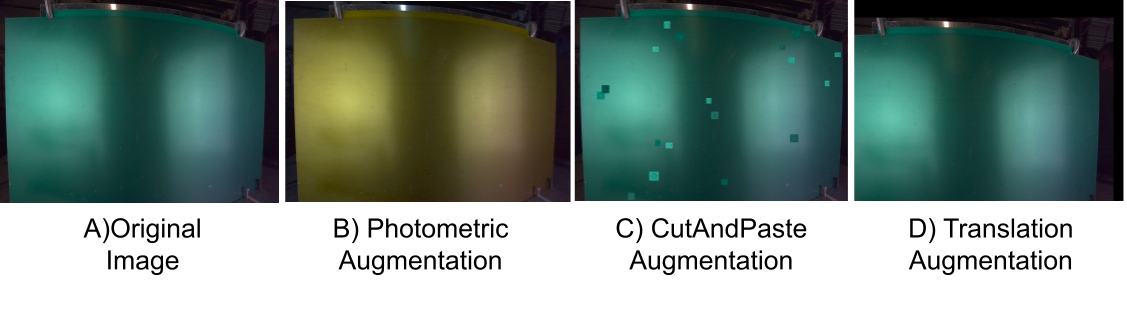}
    \caption{\textbf{Data augmentation strategies}: This visual illustrates the original image and images after a single augmentation applied to the original image. We found that these augmentations make our detection robust to illumination changes, translation variations, and clutter(bolts).}
    \label{fig:augmentation}
\end{figure}

While training, we use 3 data augmentation techniques - photometric (varying brightness, contrast, hue, and saturation), CutAndPaste \cite{ghiasi2021simple}, and translation. These augmentation techniques make our model robust to illumination changes and the presence of distracting features (like bolts and big cracks) at runtime when the inspection parts could be placed in a totally different environment and could be of different shapes. Figure \ref{fig:augmentation} illustrates the augmentation techniques applied individually. At the training time, we apply all of them at the same time. The photometric data augmentation is specifically helpful to make the model robust to lighting variation which might occur in the production environment.

\subsection{Stage II: Tactile-based defect detection}\label{sec:tactiledetect}
\begin{figure}
    \centering
    \includegraphics[width=\columnwidth]{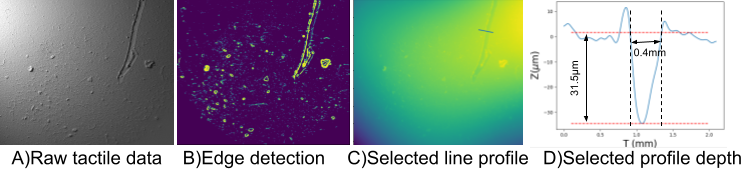}
    \caption{\textbf{Tactile detection pipeline}: The outline of our tactile sensor-based detection system A) Raw data capture by GelSight Mobile B) Output of Canny edge detection on heightfield image C) Automated anomalous profile selection D) Depth profile along the anomalous profile with width and depth annotations.}
    \label{fig:tactiledetection}
\end{figure}

We use GelSight Mobile \cite{gelsightprobe} from GelSight Inc. for obtaining high-resolution tactile information. The tactile sensor provides a high-quality heightfield as shown in Figure \ref{fig:boeingdefects} GelSight Image. Due to the high-quality heightfield, we can directly inspect anomalous regions and use the defect description to identify them. 
For figuring out the anomalous regions on heightfield, we use the canny edge detector without non-maximal suppression, followed by the Probabilistic Hough line for scratches \& drill run and Hough Circle detection for gouges, respectively. We hand-tuned the parameters of canny edge detector and feature detection algorithms. This step is required to identify potential regions containing defects. After figuring out the anomalous region, we extract the depth profile by generating a line segment passing perpendicular to the scratch \& drill run or passing through the center of the gouge, as shown in Figure \ref{fig:tactiledetection}C. After obtaining the depth profile, we detrend the depth by presuming the depth in the neighborhood of the defect is zero-level. The detrending is crucial to correctly identify the depth of the defect and use the defect definitions for identification. We use the depth and width defect descriptions, as mentioned in Table \ref{tab:defectoccurences}, for identifying the defect in the extracted profile, as shown in Figure \ref{fig:tactiledetection}. For drill run detection, we require the number of minima peaks with depth $> 10\mu m$ to be greater than 3. This heuristic is motivated by the fact that the drill run forms a repeated pattern of bumps in the specimen. 

\section{Robot system integration} \label{sec:robotsystem}
\begin{figure}
    \centering
\includegraphics[width=\columnwidth]{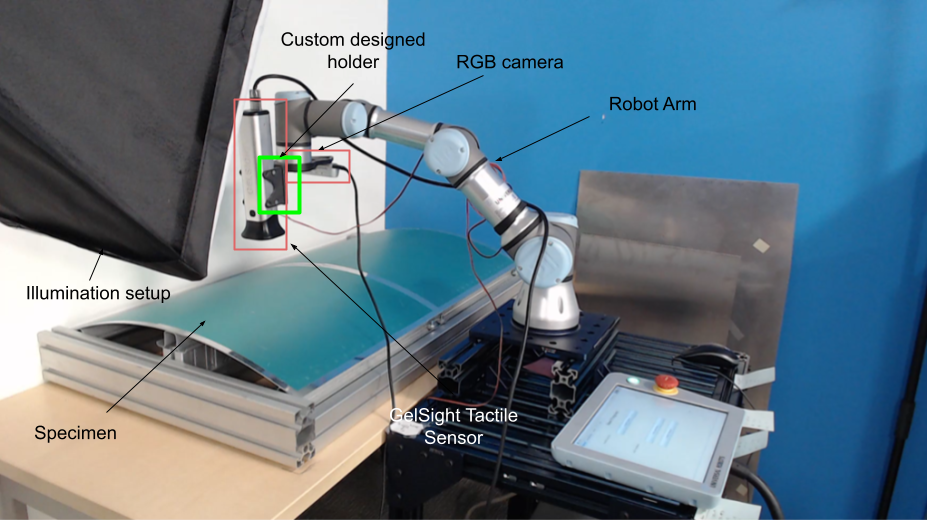}
    \caption{\textbf{Runtime System}: The robot system contains (A) UR3 robot arm (B) RealSense RGBD 435F camera (C)Neewer Illumination source (D) Custom tactile sensor mount E) GelSight Mobile 0.5x (F) Specimen under inspection. Our algorithm is run on a PC not shown in the figure.} 
    \label{fig:sysintegration}
\end{figure}

We integrated our defect detection pipeline with a robot system that is very similar to a system that can be applied for online detection in factories, as shown in Figure \ref{fig:sysintegration}.

\begin{figure}
    \centering|
\includegraphics[width=\columnwidth]{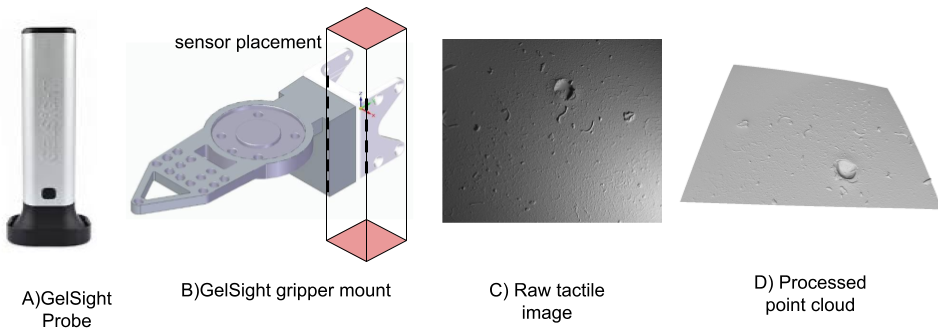}
    \caption{\textbf{GelSight tactile sensor}: A) GelSight Mobile 0.5x B) custom GelSight mount which provides compliance and good contact for high-resolution tactile image capture C) Raw RGB image captured using GelSight Mobile D) Reconstructed point cloud using GelSight Mobile} 
    \label{fig:gelsightmount}
\end{figure}

The robot system consists of a UR3 robot arm, a RealSense 435F RGBD camera mounted at the robot end-effector, a GelSight Mobile 0.5x mounted using a custom-designed mount at the robot end-effector and a Neewer 24in $\times$ 24in a softbox. Note, the depth information is not used for defect detection purposes. The robot planner and defect prediction algorithms run on a computer with Intel i7-10850H CPU @ 2.7 GHz, 6 Cores with NVIDIA Quadro T200 GPU, and Windows 10 operating system. 
The GelSight tactile sensor mount is specifically designed in order to allow compliance when indenting the metal specimen. Figure \ref{fig:gelsightmount}B shows the CAD drawing of the sensor mount. The camera to robot calibration is done using MoveIt hand-eye calibration 
. 
GelSight to end-effector transform is manually computed based on manufactured gripper mount. 

In the first stage, the robot arm collects RGB images, using the algorithm defined in Section \ref{sec:visioncontrol} and feeds them to phase I of the defect detection system described in Section \ref{sec:visiondetect}. Phase I outputs defect regions and uncertain regions. Then, the robot control uses an algorithm mentioned in Section  \ref{sec:tactilecontrol}, to collect the tactile image of each uncertain region. This tactile image is, then passed to phase II, tactile detection described in Section \ref{sec:tactiledetect}, for processing. 

\subsection{RGB Data Collection with the Robot}\label{sec:visioncontrol}
In this section, we will describe the robot control technique which is used for capturing RGB images for surface defect detection. In our current testing setup, the capture locations are pre-defined manually in the robot's task-space coordinates (3D Cartesian locations). We request that the robot collect RGB images at multiple locations to ensure the entire surface of the panel is covered. In our initial experiment, those locations are manually chosen based on the fixed position of the parts. 
The robot calculates the joint angle configuration for a task-space location using inverse kinematics \cite{murray2017mathematical}. The robot then generates joint angle trajectories toward the target joint locations using linear interpolation. We leverage the robot simulation to check for collisions and singularity. After which, the trajectory is forwarded to the robot's controller.  

\subsection{Tactile Data Collection with the Robot}
\label{sec:tactilecontrol}
In this section, we will describe the robot control strategy used to obtain tactile images using the GelSight sensor. To capture a focused tactile image, the robot needs to make the GelSight Mobile indent the surface in the perpendicular direction at the defect location. Therefore, to achieve normal indentation, we estimate coarse normal direction by obtaining a coarse depth measurement from the RGBD camera and fitting a polynomial function in $(x,y,z)$ to the specimen surface. 
Given the fitted surface function, we obtain the coarse surface normal at the target data capture location by differentiating the polynomial function w.r.t. $x$ and $y$, followed by a cross-product. We, then, use inverse kinematics and interpolation, as mentioned in the previous section, to move closer to the object. After that, we use tactile servoing until we obtain a focused tactile scan. We use background subtraction thresholding to estimate if the tactile scan is in focus. 

\section{Experiments}
To evaluate our proposed pipeline for defect detection, we perform analysis of each stage - vision only in Section \ref{sec:offlinevision} and tactile only in Section \ref{sec:offlinetactile}. We, then, perform an analysis of our two-stage inspection system integrated with a robot in Section \ref{sec:onlinerobot}. 
For our on-site robot experiments, we record the detection runtime and the accuracy of defect detection.  
\subsection{Offline Vision-based surface defect detection}\label{sec:offlinevision}
We first evaluate the performance of our vision-based algorithm for defect detection using the offline dataset introduced in Section~\ref{sec:dataset}. We fine-tuned the Neural Network using 150 training images of resolution 1280$\times$800.
We investigate the effect of using data augmentation techniques for defect detection by comparing the performance of the trained model with various augmentations. Each model was trained on 150 images for 100 iterations using SGD with a learning rate of 0.005 and weight decay of 5e-4 in PyTorch. 

\begin{figure}
    \centering
    \includegraphics[width=\columnwidth]{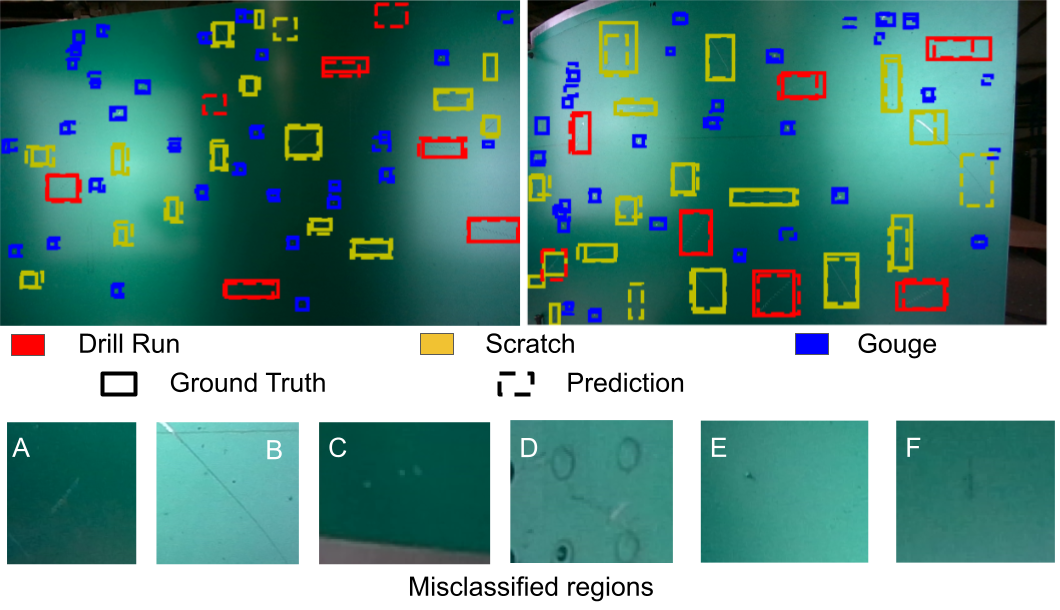}
    \caption{\textbf{RGB-only detection results in offline dataset}: We highlight the prediction of our algorithm on reserved images in our offline dataset. In the bottom row, we highlight the failure cases in detection. The common causes of failure are insufficient visual features(drill run looking like a scratch in (A)) and no depth information at the defect location(B is a paint bump instead of a scratch in the surface. The depth profile between the paint bump and scratch is significantly different).}
    \label{fig:rgbonly}
\end{figure}

During testing, we only consider bounding boxes that have a high confidence score ($0.5$ in our experiments). For calculating the recall, we used \textit{maximum detections} allowed per image to be 100. This parameter intuitively means the bounding box predictions allowed in each image. 
Table \ref{table:vision} shows the evaluation metrics using the trained Neural Network with and without augmentations. For all the metrics, we used Intersection over Union = 0.4 (metric for finding the overlap between bounding boxes) as the threshold for finding correspondence between the ground truth bounding box and the predicted bounding box. Figure \ref{fig:rgbonly} shows the test results. We found that the common misclassification cases are: (i) confusion between scratch and drill run(Figure \ref{fig:rgbonly} case A); (ii) regions that look like scratches but do not have depth(Figure \ref{fig:rgbonly} case B); (iii) very few visual features for classification(Figure \ref{fig:rgbonly} case C, D, E and F) These issues would be solved by our tactile stage, as it accounts for indentation depth and captures an orthographic view of the defect.

\begin{figure}
    \centering
    \includegraphics[width=\columnwidth]{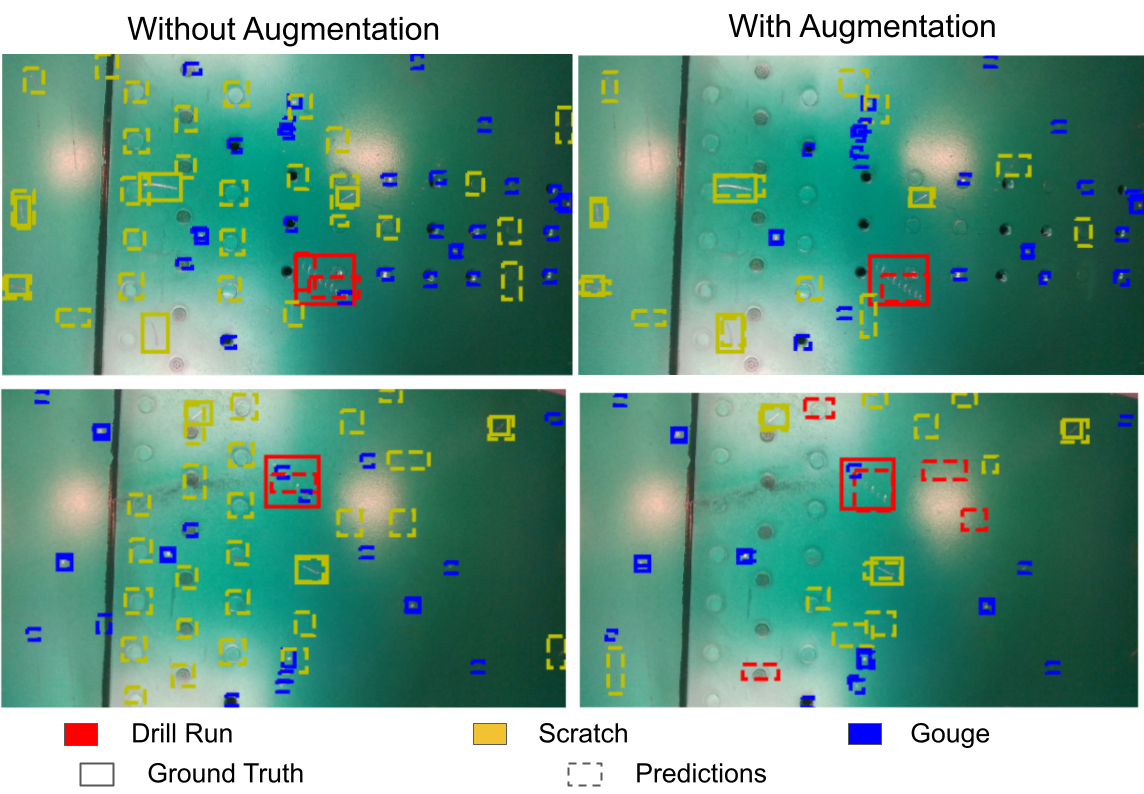}
    \caption{\textbf{Comparison of RGB-based defect detection with/without data augmentation at robot experiment time}: In this figure, the ground truth boxes are marked with solid lines, and predicted areas are marked with dashed lines. The colors of the bounding box represent \textit{drill run}, \textit{gouge}, and \textit{scratch} in red, green, and blue color respectively. The left side shows the model performance without data augmentation on 2 test images. It identifies large bolt regions as scratch defects and empty bolt regions as gouges which is incorrect. The model trained with data augmentation is able to correctly identify those regions as background as shown on the right and obtains 94.58\% recall rate without defect classification as compared to 63.56\% without augmentations.}
    \label{fig:testtimergb}
\end{figure}

In on-site robot experiments, we obtained images containing many challenging artifacts, as shown in Figure \ref{fig:testtimergb}.
Specifically, large bolt regions and bright light spots caused issues in the detection. 
Without augmentation, the probability of those areas being classified as a defect is high, as shown in Figure \ref{fig:testtimergb} left. However, with our augmentation techniques, the neural network is correctly able to identify those regions as normal regions. 

\begin{table}
\caption{Performance comparison for offline vision only defect detection system}
\scriptsize
\begin{tabular}{l|m{1.5cm}|m{1.5cm}|m{1.5cm}}
& Defect detection (without classification) & Ave. Recall with classification & Ave. Precision with classification\\
\hline
MobileNetV3+noAug &0.843 &\textbf{0.832} &\textbf{0.800} \\
MobileNetV3+allAug &\textbf{0.848} &0.829 &0.742 \\
\end{tabular}
\label{table:vision}
\end{table}

\subsection{Offline Tactile-based surface defect detection}\label{sec:offlinetactile}
\begin{figure}
    \centering
    \includegraphics[width=0.8\columnwidth]{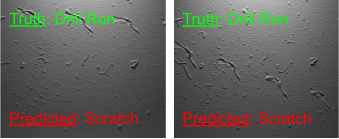}
    \caption{\textbf{Tactile detection failures}: This visual shows the illustrative failure cases in our tactile dataset with ground truth and predicted defect labels. We found 2 \textit{Drill Run} cases misclassified  because the number of repeated features was very few.}
    \label{fig:tactilemisclassified}
\end{figure}

\begin{figure}
    \centering
    \includegraphics[width=0.7\columnwidth]{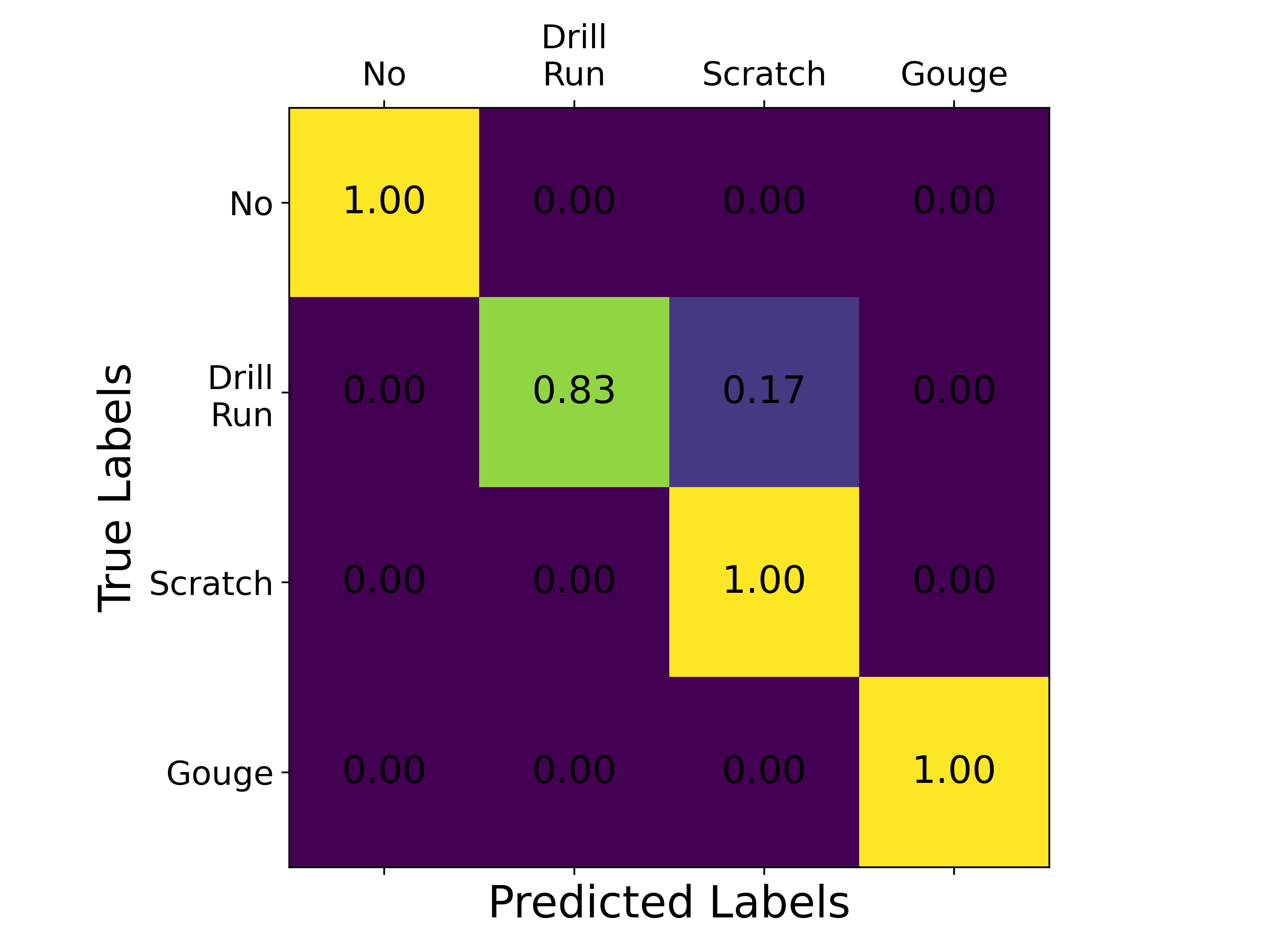}
    \caption{\textbf{Tactile confusion matrix}: We plotted the predicted label using our tactile detection algorithm on the x-axis and true labels on the y-axis. This visual highlights that our tactile detection algorithm can classify defects very well.
   }
    \label{fig:tactilecls}
\end{figure}

We evaluate the performance of the tactile-based  defect detection algorithm using the offline dataset introduced in Section~\ref{sec:dataset}.
Figure \ref{fig:tactilecls} shows the confusion matrix of the defect classification result. We obtain average classification accuracy of 95.75\%. Note, the tactile-only approach allows to identify defects with 100\% success rate if the class identification is not of concern. We notice some misclassification due to the high variability in the defects and dirt on the sensor surface in the tactile data collection. We showcase the misclassified cases in Figure \ref{fig:tactilemisclassified}. For the drill run cases, we found the depth profile is significantly different than the ideal profile according to the industrial partners and the misclassified cases have fewer drill features. Therefore, all the misclassifications are reasonable.

\subsection{Online Robot system evaluation}\label{sec:onlinerobot}
In this section, we run our integrated robotic detection system to inspect an aerospace part for potential defect regions. We capture multiple RGB images at different locations to cover the entire surface of the part. Then the tactile exploration procedure is performed on each RGB-image-covered area. 

We compare the performance of our system at runtime with vision-only and tactile-only approaches. We choose accuracy and runtime as the metric for comparison. Since tactile data capture (mean time = 22.26 seconds) takes 4x more time than visual data capture (mean time = 6.52 seconds). We use these to give an estimated time for all experiments instead of actual runtime. We use 1 panel for our robotic experiment containing 15 defects - 7 scratches, 7 gouges, and 1 Drill Run. We use 2 RGB images to cover the panel used in our experiment. Siemens engineer manually labeled the test data for this experiment. 
Table \ref{tab:runtimeperf} compares the baselines with our approach quantitatively for a new aerospace panel at Boeing's facility. Our approach achieves a perfect recall rate(@IoU=0.4 and \textit{max detections}=100) of 1.0, which is 26.5\% higher than the vision-only method and takes 0.01x of runtime as compared to the tactile-only approach. 
The defect detection system has been integrated with multiple robotic systems at 2 different locations - Siemens research lab and Boeing production labs. These environments had 2 different robotic systems - UR3 in Siemens labs and UR10 in Boeing labs. These environments had different illumination settings and panels with different curvatures for testing. This highlights that our detection is easy to adapt to various environments.

\begin{table}
    \centering
    \caption{Defect Detection and Classification Results of on-site robot experiments}
    \begin{tabular}{c||m{1.2cm}|m{1cm}|m{2cm}}
        \textbf{Method} &  
        \textbf{Average Precision} &\textbf{Average Recall} & \textbf{Runtime (seconds)}\\
        \hline
        Vision only & 0.62 & 0.79 & 13.04 \\
        Touch only &  1.0 &  1.0 & 12977.58\\
        \hline
        \textbf{Our Approach} &  1.0 &  1.0 & 168.44
    \end{tabular}
    
    \label{tab:runtimeperf}
\end{table}

\section{Conclusion}
This work introduces a robotic aerospace defect dataset and a 2-stage pipeline for defect detection on large-scale parts. Stage I uses an RGB camera to identify defect areas with a preliminary estimation, followed by the stage the robot uses a high-resolution tactile sensor GelSight Mobile for precise inspection of the potential defect area. Our approach is shown to be beneficial in terms of accuracy (perfect recall) and speed of inspection (70x faster than the tactile-only approach). We were also successfully able to integrate the detection system in 2 different environments, containing different robot arms, different illumination, and different metal panel. Comprehensive evaluation in production environment out of the scope of this research work. 

We did not have the capacity to test the robustness of the pipeline after repeated use. Touch sensor measurements become less accurate over time due to repeated interaction. Therefore, accuracy evaluations of the pipeline at repeated intervals may help the system to become robust. 
Transfer learning under significant illumination or inspection material is another avenue of research.  
Using multiple viewpoints in a single detection might be an interesting research direction to improve the accuracy of the vision stage. Another interesting extension would be to incorporate human feedback for the online update of the prediction model. 

\section{Acknowledgment}
The research is partially sponsored by Advanced Robotics for Manufacturing Institute by the Office of the Secretary of Defense and was accomplished under Agreement Number W911NF-17-3-0004. The views and conclusions contained in this document are those of the authors and should not be interpreted as representing the official policies, either expressed or implied, of the Office of the Secretary of Defense or the U.S. Government. The U.S. Government is authorized to reproduce and distribute reprints for Government purposes notwithstanding any copyright notation herein. 

\bibliographystyle{IEEEtran}
\bibliography{IEEEabrv, ref}

\begin{thebibliography}{10}
\providecommand{\url}[1]{#1}
\csname url@samestyle\endcsname
\providecommand{\newblock}{\relax}
\providecommand{\bibinfo}[2]{#2}
\providecommand{\BIBentrySTDinterwordspacing}{\spaceskip=0pt\relax}
\providecommand{\BIBentryALTinterwordstretchfactor}{4}
\providecommand{\BIBentryALTinterwordspacing}{\spaceskip=\fontdimen2\font plus
\BIBentryALTinterwordstretchfactor\fontdimen3\font minus
  \fontdimen4\font\relax}
\providecommand{\BIBforeignlanguage}[2]{{%
\expandafter\ifx\csname l@#1\endcsname\relax
\typeout{** WARNING: IEEEtran.bst: No hyphenation pattern has been}%
\typeout{** loaded for the language `#1'. Using the pattern for}%
\typeout{** the default language instead.}%
\else
\language=\csname l@#1\endcsname
\fi
#2}}
\providecommand{\BIBdecl}{\relax}
\BIBdecl

\bibitem{aopa_2016}
\BIBentryALTinterwordspacing
``Guide to aircraft inspections,'' Jul 2016. [Online]. Available:
  \url{https://www.aopa.org/go-fly/aircraft-and-ownership/maintenance-and-inspections/aircraft-inspections}
\BIBentrySTDinterwordspacing

\bibitem{nicklesdescriptive}
G.~Nickles, H.~Him, S.~Koenig, A.~Gramopadhye, and B.~Melloy, ``A descriptive
  model of aircraft inspection activities. 2019.''

\bibitem{marketsandmarkets_2020}
\BIBentryALTinterwordspacing
MarketsandMarkets, ``Surface inspection market worth \$5.3 billion by 2025 -
  exclusive report by marketsandmarkets™,'' Feb 2020. [Online]. Available:
  \url{https://www.prnewswire.com/news-releases/surface-inspection-market-worth-5-3-billion-by-2025--exclusive-report-by-marketsandmarkets-301009731.html}
\BIBentrySTDinterwordspacing

\bibitem{roth2022towards}
K.~Roth, L.~Pemula, J.~Zepeda, B.~Sch{\"o}lkopf, T.~Brox, and P.~Gehler,
  ``Towards total recall in industrial anomaly detection,'' in
  \emph{Proceedings of the IEEE/CVF Conference on Computer Vision and Pattern
  Recognition}, 2022, pp. 14\,318--14\,328.

\bibitem{jahanshahi2013innovative}
M.~R. Jahanshahi, S.~F. Masri, C.~W. Padgett, and G.~S. Sukhatme, ``An
  innovative methodology for detection and quantification of cracks through
  incorporation of depth perception,'' \emph{Machine vision and applications},
  vol.~24, no.~2, pp. 227--241, 2013.

\bibitem{weng2020multi}
T.~Weng, A.~Pallankize, Y.~Tang, O.~Kroemer, and D.~Held, ``Multi-modal
  transfer learning for grasping transparent and specular objects,'' \emph{IEEE
  Robotics and Automation Letters}, vol.~5, no.~3, pp. 3791--3798, 2020.

\bibitem{zou2018deepcrack}
Q.~Zou, Z.~Zhang, Q.~Li, X.~Qi, Q.~Wang, and S.~Wang, ``Deepcrack: Learning
  hierarchical convolutional features for crack detection,'' \emph{IEEE
  Transactions on Image Processing}, vol.~28, no.~3, pp. 1498--1512, 2018.

\bibitem{jiang2021vision}
J.~Jiang, G.~Cao, D.~F. Gomes, and S.~Luo, ``Vision-guided active tactile
  perception for crack detection and reconstruction,'' in \emph{2021 29th
  Mediterranean Conference on Control and Automation (MED)}.\hskip 1em plus
  0.5em minus 0.4em\relax IEEE, 2021, pp. 930--936.

\bibitem{palermo2021multi}
F.~Palermo, L.~Rincon-Ardila, C.~Oh, K.~Althoefer, S.~Poslad, G.~Venture, and
  I.~Farkhatdinov, ``Multi-modal robotic visual-tactile localisation and
  detection of surface cracks,'' in \emph{2021 IEEE 17th International
  Conference on Automation Science and Engineering (CASE)}.\hskip 1em plus
  0.5em minus 0.4em\relax IEEE, 2021, pp. 1806--1811.

\bibitem{bergmann2019mvtec}
P.~Bergmann, M.~Fauser, D.~Sattlegger, and C.~Steger, ``Mvtec ad--a
  comprehensive real-world dataset for unsupervised anomaly detection,'' in
  \emph{Proceedings of the IEEE/CVF conference on computer vision and pattern
  recognition}, 2019, pp. 9592--9600.

\bibitem{magnetictileHuang2020surface}
Y.~Huang, C.~Qiu, and K.~Yuan, ``Surface defect saliency of magnetic tile,''
  \emph{The Visual Computer}, vol.~36, no.~1, pp. 85--96, 2020.

\bibitem{everingham2010pascal}
M.~Everingham, L.~Van~Gool, C.~K. Williams, J.~Winn, and A.~Zisserman, ``The
  pascal visual object classes (voc) challenge,'' \emph{International journal
  of computer vision}, vol.~88, no.~2, pp. 303--338, 2010.

\bibitem{gelsightprobe}
M.~K. Johnson, F.~Cole, A.~Raj, and E.~H. Adelson, ``Microgeometry capture
  using an elastomeric sensor,'' \emph{ACM Transactions on Graphics (TOG)},
  vol.~30, no.~4, pp. 1--8, 2011.

\bibitem{fasterrcnn}
S.~Ren, K.~He, R.~Girshick, and J.~Sun, ``Faster r-cnn: Towards real-time
  object detection with region proposal networks,'' \emph{Advances in neural
  information processing systems}, vol.~28, 2015.

\bibitem{mobilenetv3}
A.~Howard, M.~Sandler, G.~Chu, L.-C. Chen, B.~Chen, M.~Tan, W.~Wang, Y.~Zhu,
  R.~Pang, V.~Vasudevan \emph{et~al.}, ``Searching for mobilenetv3,'' in
  \emph{Proceedings of the IEEE/CVF international conference on computer
  vision}, 2019, pp. 1314--1324.

\bibitem{coco}
T.-Y. Lin, M.~Maire, S.~Belongie, J.~Hays, P.~Perona, D.~Ramanan,
  P.~Doll{\'a}r, and C.~L. Zitnick, ``Microsoft coco: Common objects in
  context,'' in \emph{European conference on computer vision}.\hskip 1em plus
  0.5em minus 0.4em\relax Springer, 2014, pp. 740--755.

\bibitem{ghiasi2021simple}
G.~Ghiasi, Y.~Cui, A.~Srinivas, R.~Qian, T.-Y. Lin, E.~D. Cubuk, Q.~V. Le, and
  B.~Zoph, ``Simple copy-paste is a strong data augmentation method for
  instance segmentation,'' in \emph{Proceedings of the IEEE/CVF Conference on
  Computer Vision and Pattern Recognition}, 2021, pp. 2918--2928.

\bibitem{murray2017mathematical}
R.~M. Murray, Z.~Li, and S.~S. Sastry, \emph{A mathematical introduction to
  robotic manipulation}.\hskip 1em plus 0.5em minus 0.4em\relax CRC press,
  2017.

\end{thebibliography}

\end{document}